%% file: GUY_ICPR2020_sub.tex
\documentclass[10pt,a4paper,conference]{IEEEtran}
\usepackage{graphicx}
\usepackage{amsmath,amssymb} 
\usepackage[compress]{cite}
\usepackage{url}
\usepackage{epsfig}
\usepackage{xcolor}
\usepackage{multirow}
\usepackage{caption}
\usepackage{booktabs}
\usepackage{tabularx}
\usepackage{subfig}
\usepackage{hyperref}

\begin{document}
\title{Learning Visual Voice Activity Detection with an Automatically Annotated Dataset}
\author{\IEEEauthorblockN{
Sylvain Guy\IEEEauthorrefmark{1},
St\'{e}phane Lathuili\`{e}re\IEEEauthorrefmark{2}\IEEEauthorrefmark{1},
Pablo   Mesejo\IEEEauthorrefmark{3}\IEEEauthorrefmark{1} and
Radu Horaud\IEEEauthorrefmark{1}}
\IEEEauthorblockA{\IEEEauthorrefmark{1}  Inria  Grenoble Rh\^{o}ne-Alpes   and   Univ.  Grenoble   Alpes, France}
\IEEEauthorblockA{\IEEEauthorrefmark{2} LTCI, T\'{e}l\'{e}com Paris, Institut Polytechnique de Paris, France}
\IEEEauthorblockA{\IEEEauthorrefmark{3}  Andalusian  Research  Institute  in  Data  Science  and \\ Computational  Intelligence  (DaSCI),  University  of  Granada, Spain.}
}

\maketitle

\begin{abstract}
 Visual voice activity detection (V-VAD) uses visual features to predict whether a person is speaking or not. V-VAD is useful whenever audio VAD (A-VAD) is inefficient either because the acoustic signal is difficult to analyze or because it is simply missing. We propose two deep architectures for V-VAD, one based on facial landmarks and one based on optical flow. Moreover, available datasets, used for learning and for testing V-VAD, lack content variability. We introduce a novel methodology to automatically create and annotate very large datasets \emph{in-the-wild} -- WildVVAD --  based on combining A-VAD with face detection and tracking. A thorough empirical evaluation shows the advantage of training the proposed deep V-VAD models with this dataset.\footnote{\url{https://team.inria.fr/perception/research/vvad/}}
 \end{abstract}


\input{intro}

\input{related}

\input{model}
\input{method}

\input{experiments}

\input{conclusion}

\section*{Acknoledgments}
This work has been funded by the EU H2020 project \#871245 SPRING and by the Multidisciplinary Institute in Artificial Intelligence (MIAI) \# ANR-19-P3IA-0003.






%

\bibliographystyle{IEEEtran}


\end{document}

%% file: intro.tex
\section{Introduction}
\label{sec:intro}

Voice activity detection (VAD) is of great importance in auditory, visual or audio-visual scene analysis. A-VAD refers to VAD solely based on audio signals, while
visual V-VAD uses visual information. A-VAD has already been studied for many years \cite{Ramirez2007-VAD}. 
 Its main limitation 
 is when competing speech or non-speech signals are simultaneously present. By opposition to A-VAD, V-VAD is insensitive to other acoustic sources such as background noise, and to mixed speech signals since each potential speaker can be analyzed independently. 
For these reasons, V-VAD is extremely useful whenever the audio signals are too complex to be analyzed, or when the speech uttered by a person is not available at all.

Deep neural networks trained on large datasets have led to impressive results in visual analysis tasks, 
 but in the case of V-VAD, current publicly available datasets have a small size and/or do not usually include scenes under various lighting conditions, head poses, video quality, or background. This greatly limits the possibilities of progress in this research field.

The paper contribution is manyfold. We introduce two DNN models, a landmark-based model and an optical-flow-based model, and we show that both these models outperform state-of-the-art V-VAD, whether landmark- or optical-flow-based.
We propose a method that automatically generates and annotates in-the-wild V-VAD datasets, namely speaking or silent faces with natural head motions and facial expressions. We show that V-VAD trained with these datasets perform well in the presence of complex social situations. Whether automatic or manual, annotation errors are inherently present. Our method generates a dataset with an annotation error of approximately 10\%. We show that this error affects the proposed V-VAD methods moderately. We compare the results obtained with the automatic annotation (possibly noisy) of a large dataset with the results obtained with the manually cleaned annotation of a small dataset and we show that V-VAD trained with the former dataset outperforms V-VAD trained with the latter dataset.

%% file: related.tex


\section{Related Work}
\label{sec:related_work}

Two main methodological families exist for V-VAD \cite{autospeechreco}. First, methods based color, texture, or optical flow extracted from the mouth region of interest (ROI), e.g.  \cite{vadvisualinfo,inthewild,mouthintensities}. Second, there are shape-based approaches which exploit features such as the lip contour, tongue or teeth positions, e.g. \cite{Liu2014ITM,analysisVisual}. Other approaches rely on combining several features, often based on their concatenation, e.g.\cite{appearenceandretine,vvadadaboost}. These methods are currently evaluated on datasets that are not publicly available, on small datasets, or on datasets with weak inter-sample variabilities, e.g. high-resolution videos of participants facing the camera.


DNNs have become the state-of-the-art for many classification tasks in computer vision. However, until recently, DNN-based methods for V-VAD are scarce. To the best of our knowledge, they were mainly used in constrained setups, e.g. frontal views, such that good lip detection is reliably performed.

 Patrona et al. \cite{inthewild} proposed to train an appearance-based V-VAD using a dataset collected in-the-wild from three movies. Their approach uses handcrafted spatio temporal features.
 A bag-of-words (BoW) representation is employed before classifying as speaking or silent. Also, other appearance-based approaches \cite{sharma2019toward} employ commercial movies to extract videos. However, the in-the-wild feature of these carefully recorded videos is questionable. 



 \begin{figure*}
 \centering
 \begin{tabular}{cc}
 \subfloat[Land-LSTM]  { \includegraphics[width=0.48\linewidth]{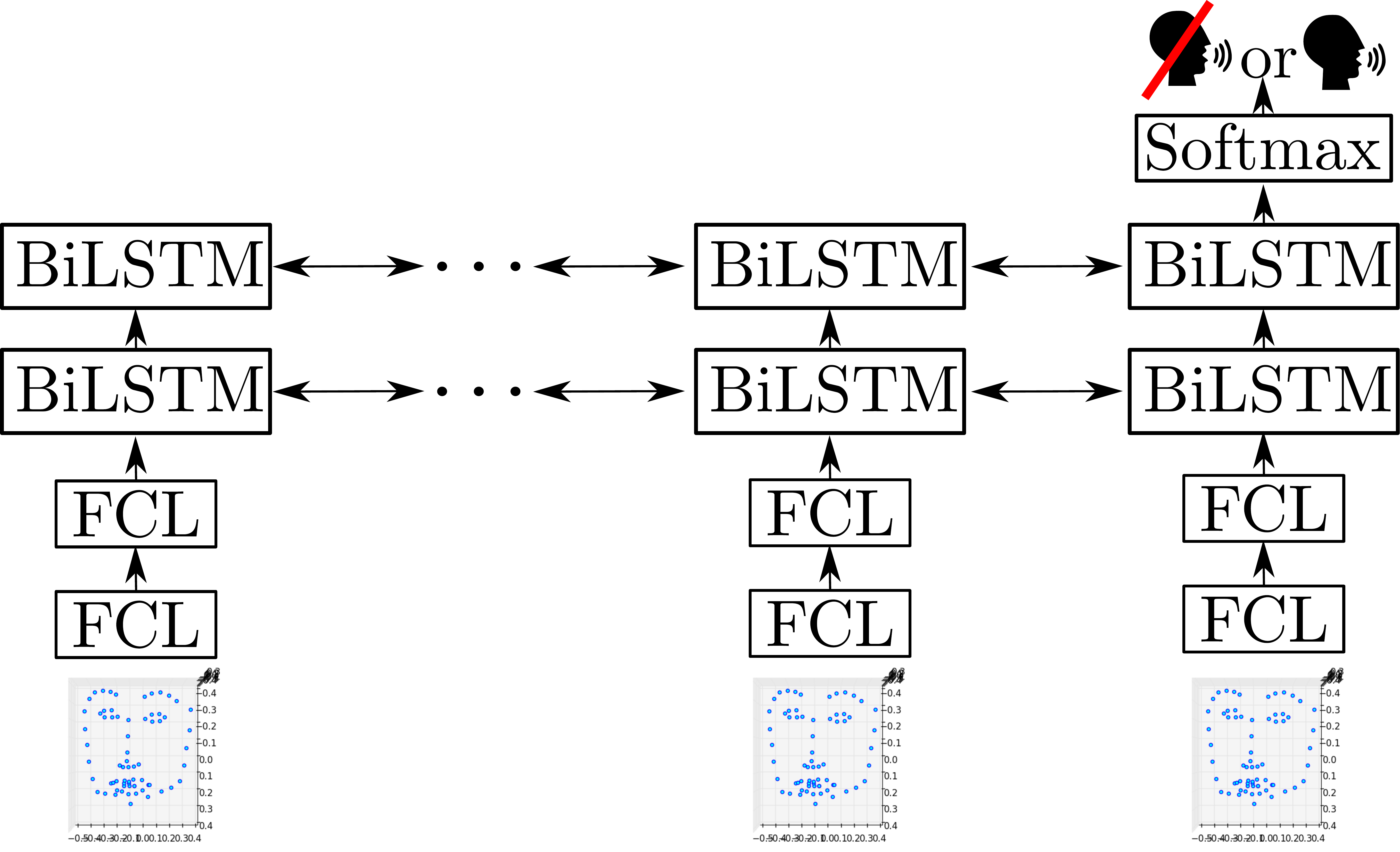}}&
 \subfloat[OF-ConvNet] {\includegraphics[width=0.48\linewidth]{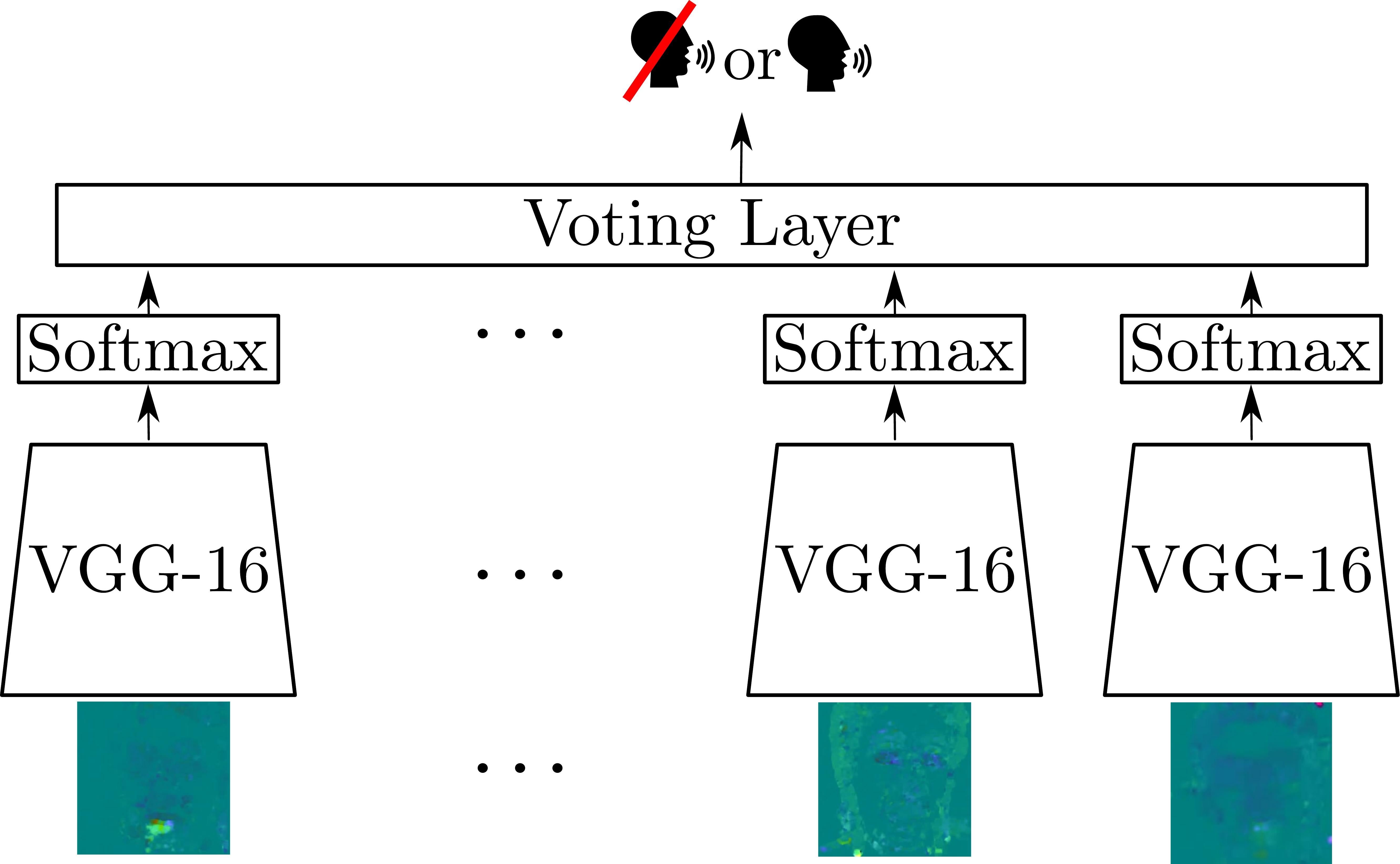}}
 \end{tabular}
 \caption{\label{fig:architectures} Architectures of the two proposed models. Both networks take as input a sequence of frames and predict as output an activity label, \emph{speaking} or \emph{silent}.}
 \end{figure*}

Several datasets for training and testing V-VAD have already been proposed.
The first datasets were limited to a small number of speakers and presented low diversity in terms of head poses and background. 
CUAVE \cite{cuave} is one among these datasets, publicly available, containing 7000 utterances of 36 male/female speakers in frontal and profile views. CUAVE has been  frequently used, even if it was not purposively built for VAD.
The dataset of \cite{biHMM} was recorded in less constrained conditions but 
is not publicly available.
In \cite{Minotto2014} a dataset for multimodal VAD, namely MVAD was proposed. This dataset is composed of 24 recordings (depth, color, and audio) lasting between 40 and 60 seconds. 
One of the first in-the-wild datasets, \cite{inthewild}, contains 4194 videos extracted from three movies with 126 actors. This dataset is not publicly available.


Other widely used datasets, e.g.~GRID \cite{grid}, are not better suited, either because they contain too few training samples with poor variability, or because they are recorded in very constrained environments. 
 The automatic generation and annotation of large corpora has become an important topic in computer vision 
 with the advent of DNNs. Indeed, DNN techniques require quite large datasets, and manual (or partially manual) annotation can be extremely tedious, time-consuming and error-prone. 
 For all these reasons, we propose a methodology to (i) generate a dataset by data harvesting, thus guaranteeing to contain very different participants, head orientations, facial features, all at various resolutions and in complex backgrounds and lighting conditions and (ii) to automatically annotate it using computer vision and audio signal processing algorithms.

%% file: model.tex
\section{Deep Models for V-VAD}
\label{sec:deep}

We describe two DNN V-VAD architectures, e.g. Figure~\ref{fig:architectures}. We propose a landmark-based model and an optical-flow-based model, in order to analyze their potential. While the former is based on sparse landmarks, the latter uses dense motion features. We aim at comparing these fundamentally different methods depending on the training and test settings. V-VAD is formulated as a classification problem: short videos of human faces are labeled either speaking or silent. 

\subsection{Combining Facial Landmarks with LSTM}
We propose a method based on 3D facial landmarks \cite{bulat2017far}. Their method returns the 3D coordinates of 68 facial landmarks in, where the $x$ and $y$ coordinates correspond to the image plane and the $z$ coordinate corresponds to depth. We perform a two-stage data preprocessing in order to obtain resolution- and pose-invariant facial descriptions. First, we normalize the size using the Euclidean distance between the landmarks corresponding to the outmost eye corners. Second, we optimally estimate the rigid transformation (rotation matrix and translation vector) that align the 68 landmarks of a face with the 68 landmarks corresponding to a frontal mean face. In that way, all the facial landmarks are brought into a frontal pose and the corresponding facial descriptor is invariant to rigid head movements. Each video sample is therefore mapped onto a sequence of vectors, where each vector contains the frontal 3D coordinates of the 68 landmarks. This sequence is then fed into a bidirectional LSTM,\cite{graves2005framewise},
in which fully connected layers share the parameters across time. 
We employ \emph{ReLu} activations and add batch normalization layers \cite{ioffe2015batch} between bidirectional LSTM layers. This method is referred to as \emph{Land-LSTM} is the rest of the paper.

\subsection{Exploiting Optical Flow with CNN}

Inspired by \cite{cheron2015p}, we propose a second method that consists of applying a CNN to the optical flow represented as an RGB image. More precisely, we first compute the optical flow 
 in all videos and 
convert it into RGB representations. The color encodes the norm and the direction of the flow. We then follow the common approach that consists of using a model pre-trained on ImageNet \cite{ILSVRC15} and of fine-tuning the last layers as proposed in \cite{cheron2015p}. We chose the widely used VGG-16 network \cite{vgg}. The
speaking status is predicted by combining the frame-wise decisions according to a voting scheme. However the model is trained without the voting layer by minimizing the cross-entropy loss for each individual frame. This method is referred to as \emph{OF-ConvNet} is the rest of the paper.

%% file: method.tex
\begin{figure*}[t!]
\centering
 \includegraphics[width=0.85\textwidth]{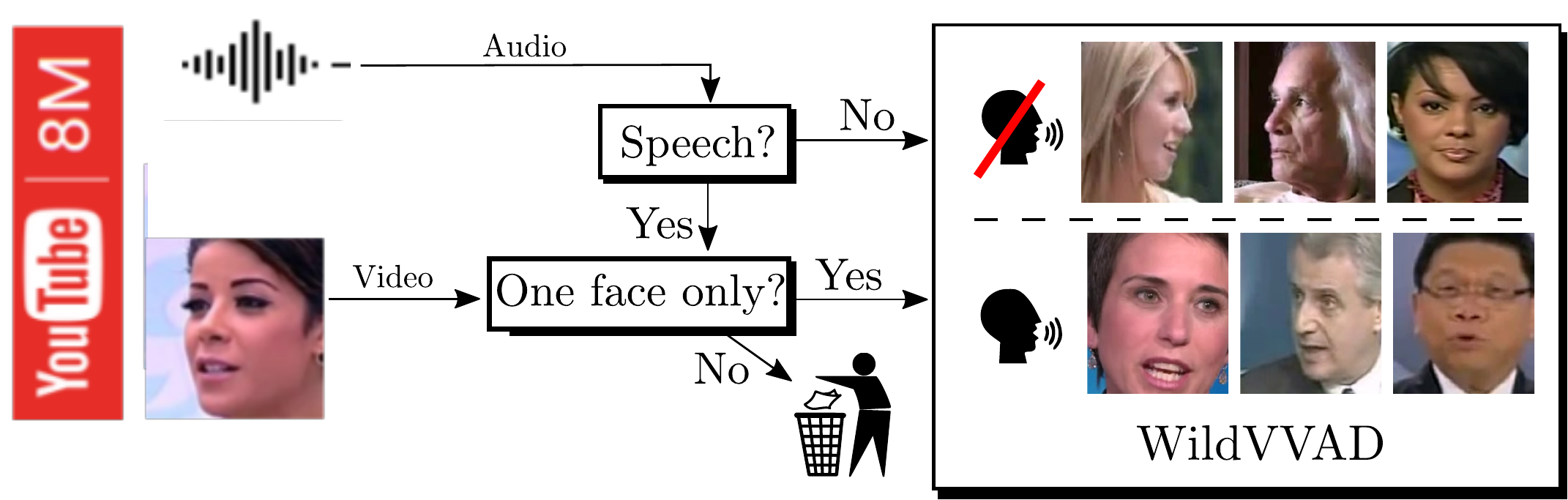}
\caption{Face detection and audio voice activity detection (A-VAD) are applied to each video sequence. Speaking faces are generated whenever A-VAD detects speech in conjunction with a single face. Silent faces are generated whenever A-VAD does not detect any speech.}
\label{fig:teaser}
\end{figure*}

\section{Automatic Dataset Generation and Annotation}
\label{sec:datasetGeneration}
\label{subsec:pipeline}

The proposed pipeline is shown in Figure \ref{fig:teaser}
and it is based on the following strategy. If there is no speech detected in the audio track, then all the faces in the video are labeled \emph{silent}. Otherwise, if speech is detected in the audio track, two cases are considered depending on the number of faces that are present in the image sequence. If a single face is detected, it is labeled 
\emph{speaking}. If several faces are present, the corresponding image sequence is discarded from the dataset. This methodology simply requires A-VAD and face detection.
One should be aware that this strategy generates false positives whenever the audio signal doesn't match the face, such as a sound uttered by someone outside the field of view. Below we provide details of each step of the proposed method.

\subsection{Input Videos} In order to build a large dataset, a huge amount of input videos is required. The YouTube8M dataset \cite{youtube8m} is particularly well suited for this task since it contains several millions of videos and it provides playlists of videos tagged with a class label. However, not all classes are equally suited for extracting faces, so shortlisting a few labels is necessary. As mentioned above, a key assumption of our approach is that if speech and a single face are simultaneously detected, the face is labeled speaking. However and as already mentioned, one should avoid false positives. To do this, we selected playlists mostly containing interviews and TV shows. More precisely, we downloaded videos from the BBC collection (3225 videos), Newscaster collection (11865), and Television collection (12225 videos). The selection of these three lists is the only supervision of the proposed automatic dataset building strategy. These videos are further split into short (one to two seconds) samples by detecting sudden changes between two consecutive frames.

\subsection{Audio Voice Activity Detection} We used WebRTC \cite{recentvvad} to perform A-VAD. WebRTC employs multiple-frequency (subband) features combined with a pre-trained GMM classifier. This is much more effective than a simple energy-threshold detector, particularly in the presence of dynamic types and levels of background noise. 

\subsection{Face Detection and Tracking} The next step is to extract faces and to track each face over time. The face detector should not provide false positives as it would introduce mislabeled videos in the dataset. In turn, false negatives are not a major issue since they would just imply that some faces are ignored. 
 Several face detectors were tested, and a DNN-based method \cite{dlibcnn} was finally chosen for its low false positives rate.

To perform face tracking, we go through the sequences of detected faces in order to associate them to a face index, e.g. an anonymous identity. Our goal is to ensure that only one identity appears on each track: if some tracks are wrongly lost, it would, again, only increase the amount of downloaded videos needed to generate the dataset. Given this particular context, first, we use non-maxima suppression 
 to remove overlapping detections; second, we assign indexes based on the distance between the bounding box centers and the previous centers. A new track is created if the distance to all previous bounding boxes is greater than a threshold.

The tracks correspond to a sequence of bounding boxes. We apply a Kalman smoother to the bounding-box trajectories. Finally the 
bounding-box dimensions are extended to the largest bounding box dimension to obtain bounding boxes of the same size. When the bounding-box coordinates exceed image boundaries, replicate-padding is employed.

\subsection{Quantitative and Qualitative Assessment of WildVVAD}
\label{subsec:evaldata}

The WildVVAD dataset that we built contains 13,000 two-second videos. The quantitative evaluation has to be considered from two different perspectives. First, the pertinency of the proposed pipeline has to be measured, i.e. the amount of incorrectly labeled videos in the dataset produced by the method needs to be assessed. This is performed by carefully inspecting a subset of the videos in the dataset, thus revealing $12\%$ and $8.6\%$ mislabeled speaking and silent videos, respectively. We empirically show in Section~\ref{expe} that this percentage of mislabeled samples is not a major issue in order to effectively learn V-VAD models. Second, we have to measure the performance yielded by a classifier trained on the dataset. In this case we need a clean test set to precisely evaluate the performance. For these two reasons, we randomly selected 978 two-second videos which were subsequently manually annotated, leading to an error-free test set composed of 441 speaking samples and 537 silent samples. It is important to remark that the manual annotation of this small data subset is only necessary for evaluation purposes.  


\begin{figure*}[h!]
\centering
\begin{tabular}{c}
 \includegraphics[width=0.09\textwidth]{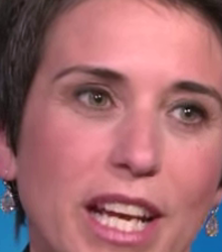}
\includegraphics[width=0.09\textwidth]{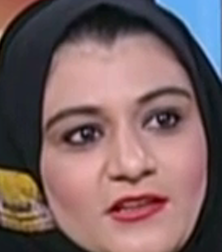}
\includegraphics[width=0.09\textwidth]{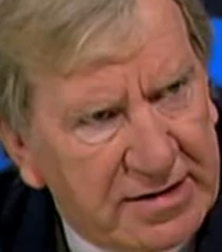}
\includegraphics[width=0.09\textwidth]{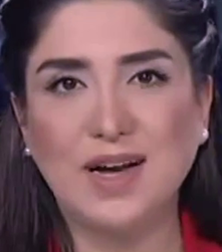}
\includegraphics[width=0.09\textwidth]{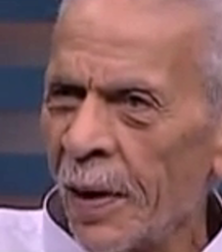}
\includegraphics[width=0.09\textwidth]{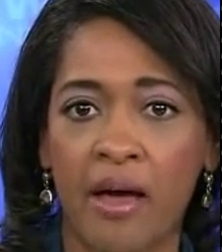}
\includegraphics[width=0.09\textwidth]{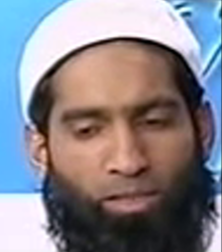}
\includegraphics[width=0.09\textwidth]{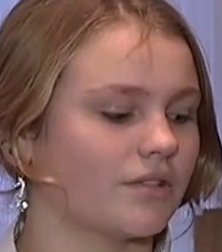}
\includegraphics[width=0.09\textwidth]{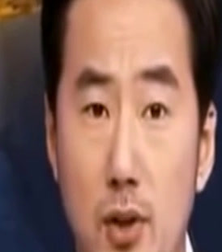}
\includegraphics[width=0.09\textwidth]{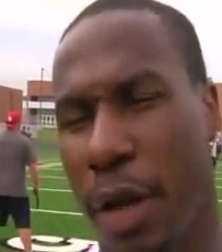}\\
\includegraphics[width=0.09\textwidth]{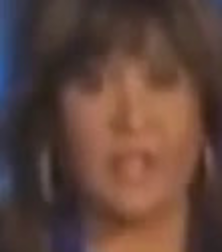}
\includegraphics[width=0.09\textwidth]{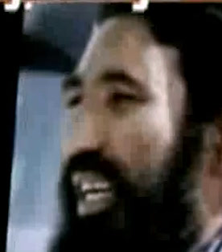}
\includegraphics[width=0.09\textwidth]{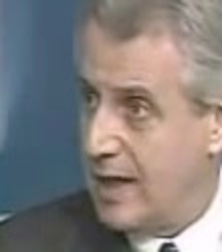}
\includegraphics[width=0.09\textwidth]{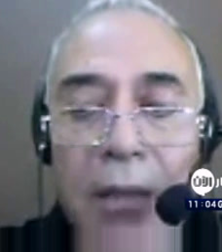}
\includegraphics[width=0.09\textwidth]{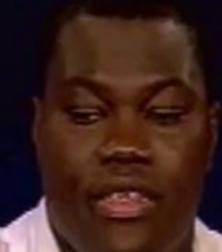}
\includegraphics[width=0.09\textwidth]{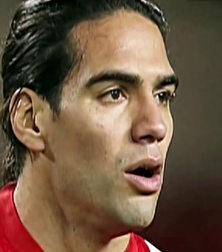}
\includegraphics[width=0.09\textwidth]{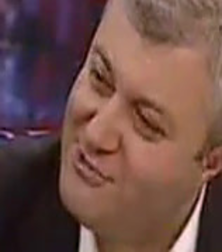}
\includegraphics[width=0.09\textwidth]{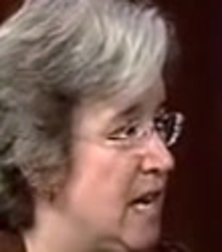}
\includegraphics[width=0.09\textwidth]{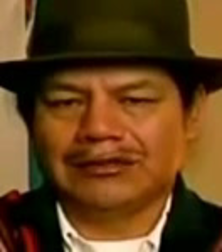}
\includegraphics[width=0.09\textwidth]{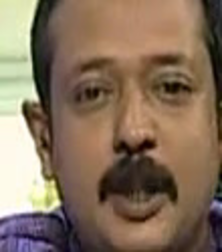}\\
(a) Speaking examples \\
\includegraphics[width=0.09\textwidth]{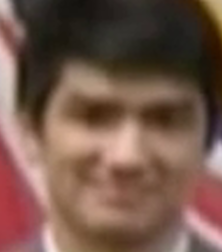}
 \includegraphics[width=0.09\textwidth]{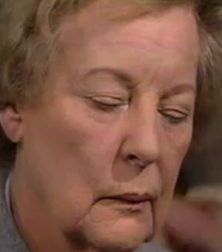}
\includegraphics[width=0.09\textwidth]{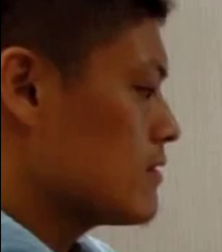}
\includegraphics[width=0.09\textwidth]{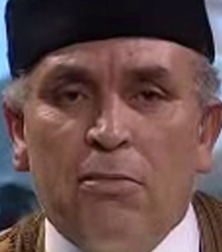}
\includegraphics[width=0.09\textwidth]{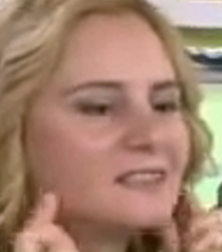}
\includegraphics[width=0.09\textwidth]{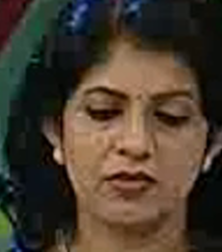}
 \includegraphics[width=0.09\textwidth]{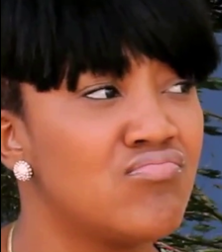}
\includegraphics[width=0.09\textwidth]{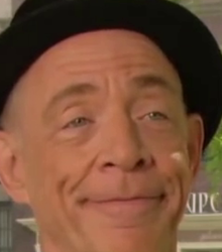}
\includegraphics[width=0.09\textwidth]{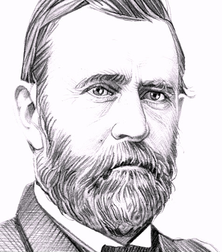}
\includegraphics[width=0.09\textwidth]{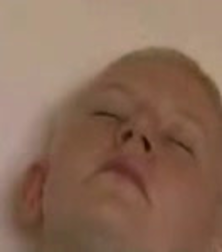}\\
\includegraphics[width=0.09\textwidth]{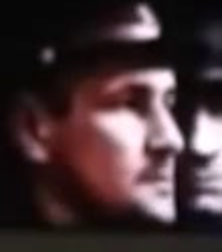}
\includegraphics[width=0.09\textwidth]{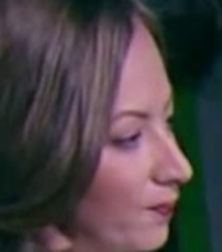}
\includegraphics[width=0.09\textwidth]{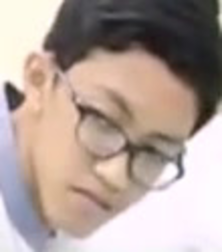}
\includegraphics[width=0.09\textwidth]{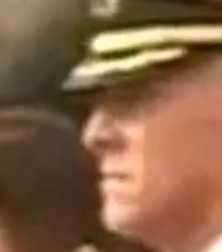}
\includegraphics[width=0.09\textwidth]{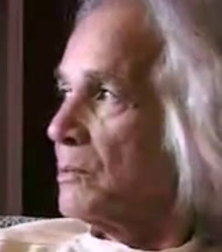}
\includegraphics[width=0.09\textwidth]{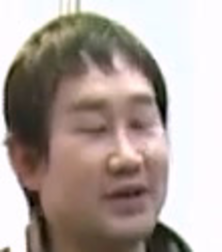}
\includegraphics[width=0.09\textwidth]{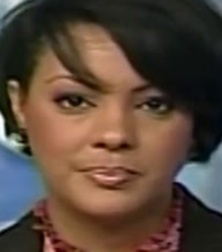}
\includegraphics[width=0.09\textwidth]{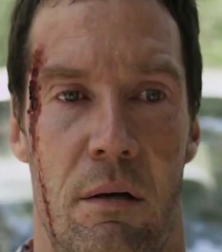}
\includegraphics[width=0.09\textwidth]{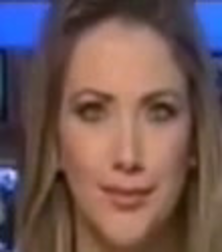}
\includegraphics[width=0.09\textwidth]{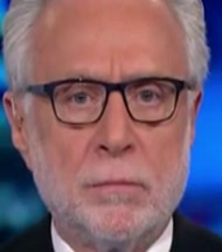}\\
(b) Silent examples
\end{tabular}
\caption{Speaking and silent examples that were automatically generated in order to build the WildVVAD dataset. 
}
\label{fig:examples_in_dataset}
\end{figure*}

\begin{figure}[h]
\centering
\includegraphics[width=0.115\textwidth]{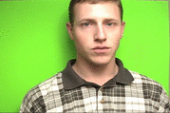}
 \includegraphics[width=0.115\textwidth]{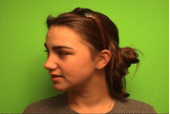}
\includegraphics[width=0.115\textwidth]{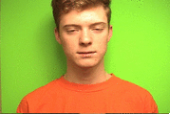}
\includegraphics[width=0.115\textwidth]{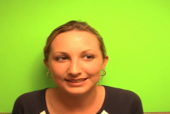}\\
\includegraphics[width=0.115\textwidth]{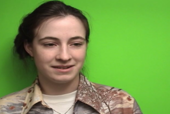}
\includegraphics[width=0.115\textwidth]{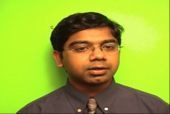}
\includegraphics[width=0.115\textwidth]{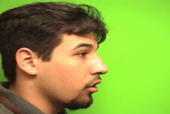}
\includegraphics[width=0.115\textwidth]{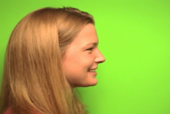}
\caption{Examples from the CUAVE dataset.} 
\label{fig:examplesCUAVE}
\end{figure}

From a qualitative point of view, as illustrated in Figure~\ref{fig:examples_in_dataset}, this dataset is extremely diverse, containing very different head poses, resolutions (most of them are around 158x158 pixels) and video quality, various types of people (in terms ethnicity, age, gender, etc.), different lighting conditions, and a large variability in lip movements. 

%% file: experiments.tex
\section{Experiments}
\label{expe}

\begin{table}[b!]
    \centering
   \caption{Results obtained on CUAVE and WildVVAD.}
\begin{tabular}{lccc}
  \toprule
 Method & TPR$\pm \sigma$ & TNR$\pm\sigma$  & ACC$\pm\sigma$ \\
  \toprule
\multicolumn{4}{c}{  \textbf{CUAVE}}\\
 \midrule
 \cite{siatras2006visual} & - &- & $52.8\%$\\
      \cite{navarathna2010dynamic} &- &- & $71.3\%$\\ 
      \cite{navarathna2011visual} & -& -& $74.1\%$\\ 
    \emph{STIP} \cite{stipimplem} & $98.40 \pm 2.19\%$  & $77.78 \pm 4.40\%$  &${\bf 89.81} \pm 1.97\%$\\
    \emph{DTs} \cite{densetrajimplem} & $96.02 \pm 4.01\%$  & $67.80 \pm 15.04\%$  &$84.26 \pm 6.92\%$\\
    \emph{OF-ConvNet} & $\bf 100.0 \pm0.00\%$ & $15.48 \pm 1.37\%$ & $64.52 \pm0.60\%$\\
  \emph{Land-LSTM} & $89.60 \pm 4.63\%$ & ${\bf83.02} \pm 14.63\%$& $84.31 \pm 9.41\%$\\
  \toprule
\multicolumn{4}{c}{\textbf{WildVVAD}} \\
  \toprule
\emph{STIP} \cite{stipimplem}& $84.80\%$ &$ 71.00\%$ &$77.90\%$\\
    \emph{DTs} \cite{densetrajimplem} & $82.10\% $& $74.00\%$ &$78.05\%$\\
       \emph{OF-ConvNet} & $81.18\%$ & $\bf 88.55\%$ & $85.38\%$\\
  \emph{Land-LSTM} & $\bf 91.26\%$ & $80.73\%$ &$\bf 86.21\%$\\
      \bottomrule
  \end{tabular}
  \label{table:results}
\end{table}

Classical computer vision techniques and deep learning approaches are compared. 
 First, each model is evaluated on the CUAVE dataset and on WildVVAD. From this comparison, we validate the two proposed V-VAD methods and evaluate them using the CUAVE and WildVVAD datasets. This enables us to compare the quality of the trained models with respect to two different datasets: an in-the-wild dataset that contains annotation errors and a dataset recorded in a constrained environment with no annotation error. Moreover, we perform a cross-dataset experiment (training with one dataset and testing with another one).
 
\subsection{Experimental Protocol}

The CUAVE dataset, e.g. Figure~\ref{fig:examplesCUAVE}, contains 36 five-minute videos of 36 different speakers. We automatically annotated 434 videos (181 silent samples and 253 speaking samples) using the associated timestamps and video scripts. There are no silent video annotations. We adopted a strategy, also used by other authors, that consists of detecting pauses between consecutive speech intervals and annotating them as silent sequences. Among all available videos in the dataset, the ones including frontal and profile faces were selected, as commonly done in the  literature. In total, this version of the CUAVE dataset includes 181 videos and 253 videos of silent and speaking people, respectively. It also includes videos with people moving while speaking.

We use the following metrics: the true positive rate (TPR) which computes the ratio of videos correctly classified as speaking over the total speaking videos, the true negative rate (TNR) which computes the same ration but for silent videos, and the accuracy (ACC). In CUAVE, we use 5-fold cross-validation, while on WildVVAD we use holdout where the manually annotated samples are used for test.    
Concerning the \emph{OF-ConvNet} model, we fine-tune the last two convolutional blocks of VGG-16. For both \emph{OF-ConvNet} and \emph{Land-LSTM}, we employ the adam optimizer \cite{kingma2014adam}. Following \cite{lathuiliere2019comprehensive}, for both datasets, we use $20\%$ of the training dataset as a validation set to perform early stopping with a patience of 7 epochs, meaning that we stop training after 7 consecutive epochs without reducing the loss value of the validation set.  For both architectures, the batch size is set to 64.

\subsection{Handcrafted method}
For the sake of comparison, we re-implemented the method described \cite{inthewild} on the following grounds. Two feature extraction techniques are tested. The first feature extraction technique is the computation of HOG and HOF descriptors on STIP using the implementation of \cite{stipimplem}. In this implementation, the descriptors are computed on a 3D video patch in the neighborhood of each detected STIP. This patch is partitioned into a grid with 3x3x2 spatio-temporal blocks, and 4-bin HOG  and 5-bin HOF descriptors are computed for all blocks. All these descriptors are then concatenated into a 72-element and a 90-element vector, respectively. All parameters were set to their default values. 

The second feature extraction technique is the computation of dense trajectories (DTs) \cite{densetrajimplem}. In this case, HOG, HOF, trajectories and MBH features were computed from the detected keypoints using the  default parameter values. In both cases, all  descriptors are then $L_{2}$ normalized, concatenated, and a K-means algorithm is used to cluster the data in a codebook of size 2000, as in \cite{inthewild}. A bag of words (BoW) technique is applied using the fuzzy quantization algorithm explained in the same paper. The BoW representation of videos is normalized and then trained and tested employing a kELM classifier. Importantly, as mentioned in Section~\ref{sec:related_work}, many videos are too short or too static to provide any keypoint. Following the experimental protocol of \cite{inthewild}, when this happens, these videos are automatically classified as silent. 

\subsection{Experimental Evaluation}
\label{methodCompa}
For comparison, we re-implemented the method described \cite{inthewild}.
Table~\ref{table:results} shows the results obtained on
the two datasets. For CUAVE, $\sigma$ refers to the standard deviation of the score obtained for each metric for the five evaluated folds. Holdout has been used in WildVVAD (only one value is displayed).  
 
Concerning the \emph{OF-ConvNet} approach, we notice that it does not
perform well with the CUAVE dataset whereas it outperforms handcrafted
feature-based methods with WildVVAD. It suggests that methods based on
handcrafted features perform well on small and constrained datasets,
whereas \emph{OF-ConvNet} and \emph{Land-LSTM} perform better in more complex settings, if the
training dataset is sufficiently large and diverse. We suspect that the poor results obtained by the \emph{OF-ConvNet} for classifying silent videos (low TNR) can be explained by the small size of this dataset. Indeed, the very deep structure of the VGG-16 architecture employed in \emph{OF-ConvNet} implies a large number of parameters to be estimated, compared to the \emph{Land-LSTM} model. Consequently, it may require more training data. Conversely,  \emph{Land-LSTM} performs well on both datasets. Despite the small training-set size of CUAVE, it achieves an accuracy similar to the handcrafted DTs method \cite{densetrajimplem}. The two proposed deep methods perform similarly on WildVVAD.

\subsection{Generalization Study}
We now compare the different methods trained on several datasets that lead good generalization. In other words, we evaluate how the learned representations can be transferred across datasets. This evaluation is carried out with two goals in mind. First, we want evaluate which methods are able to generalize better across datasets. Second, we measure the impact of the source dataset on which the model is trained. All methods are trained either on CUAVE or WildVVAD, and tested on the MVAD dataset \cite{Minotto2014}. When training on CUAVE, we employ the split of the first fold that is composed of a training and test sets. As we do not need the test set in this experiment, we use the test set as a validation set for early stopping. In WildVVAD, we use the training/validation sets employed in Section \ref{methodCompa}.

\begin{figure}[h]
\centering
\subfloat[Speaking examples]{
\includegraphics[width=0.24\columnwidth]{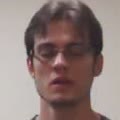}
\includegraphics[width=0.24\columnwidth]{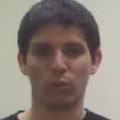}
\includegraphics[width=0.24\columnwidth]{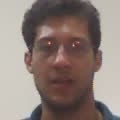}
\includegraphics[width=0.24\columnwidth]{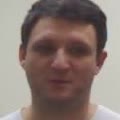}}\\
\subfloat[Silent examples]{
\includegraphics[width=0.24\columnwidth]{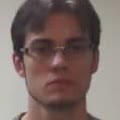}
\includegraphics[width=0.24\columnwidth]{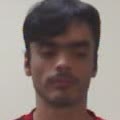}
\includegraphics[width=0.24\columnwidth]{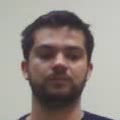}
\includegraphics[width=0.24\columnwidth]{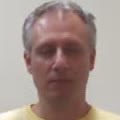}}
\caption{Examples of video frames from the MVAD dataset after applying subsequence extraction. These examples illustrate the difficulties caused by the low image resolution/quality.} 
\label{fig:examplesMVAD}
\end{figure}

This MVAD dataset consists of 24 video sequences, each lasting
from 40 to 60 seconds. Each video contains one to three persons that randomly alternate between speaking and being silent. 
The videos were recorded using two different cameras leading to two different image resolutions.
MVAD is only used as a test dataset, in order to quantify the ability of the proposed models, trained with WildVVAD and CUAVE, to perform well when tested with MVAD.
One prominent difficulty of this dataset, in our context, is that both the image resolution and the image quality are lower than on the training datasets. For each person present in a video, we extract all the 50 frame-long subsequences whenever the speaking state is constant. We obtain a test set composed of 356 sequences. Figure \ref{fig:examplesMVAD} shows some example images from the MVAD dataset after preprocessing. 

\begin{table}[h!t!]
\centering
   \caption{Results obtained on the MVAD dataset when training on CUAVE and WildVVAD.}
\begin{tabular}{llccc}
  \toprule
\multicolumn{4}{c}{  \textbf{Trained on CUAVE}}\\
  \toprule
   Method & TPR & TNR  & ACC \\
  \midrule
   \emph{STIP} \cite{stipimplem} &76.74\%  & 34.24\% &54.78\%\\
  \emph{Land-LSTM} & {\bf 81.42\%} & {\bf 59.79\%} & {\bf 64.0\%}\\
  \toprule
\multicolumn{4}{c}{\textbf{Trained on WildVVAD}} \\
  \toprule
   Method & TPR & TNR  & ACC \\
  \midrule
\emph{STIP} \cite{stipimplem} & 21.44\%  & 76.65\%  &49.98\%\\
       \emph{OF-ConvNet} &$62.69\%$  &$68.71\%$  & $65.45\%$ \\
  \emph{Land-LSTM} & $\bf 91.30\%$& $\bf 90.06\%$ & $\bf 91.01\%$ \\
  \midrule
  \end{tabular}
\label{table:resultsMVAD}
\end{table}

Table \ref{table:resultsMVAD} reports the results obtained with each method when training on CUAVE and on WildVVAD. 
Results using DTs are note shown as they systematically predict the speaking class only in this case.
\emph{Land-LSTM} performs well when tested with MVAD. WildVVAD training achieves higher test scores than CUAVE training. The accuracy is higher than the accuracy reported in \cite{Minotto2014} ($85.55\%$) where MVAD is used for both training and testing. 
 Finally, we observe that (\emph{OF-ConvNet}), which obtains good performances, e.g. Table \ref{table:results}, has a less good generalization ability than \emph{Land-LSTM}. This can be explained by the fact that shape description using facial landmarks has interesting photometric-invariant properties. We conclude that facial landmarks coupled with LSTM is the method of choice.
 

%% file: conclusion.tex
\section{Conclusions}
\label{sec:conclusions}

The contribution of the paper is twofold. We proposed two deep architectures, one based LSTM using facial landmarks and one based CNN using optical flow. We also proposed a methodology for the full automatic generation and annotation of a large in-the-wild dataset. Compared with existing V-VAD datasets, we showed that the proposed one is better suited for training V-VAD as it contains speaking and silent faces with large variabilities. We conducted an empirical evaluation that has shown that the landmark-based method, when trained with the new dataset, outperforms all the other methods. 
As future work, we plan to integrate our V-VAD methods into human-robot interaction scenarios such as in \cite{lathuiliere2019neural}.